\documentclass[acmsmall]{acmart}

\usepackage{subfigure}
\usepackage{multirow}
\usepackage{makecell}
\usepackage[figuresright]{rotating}
\usepackage{wrapfig}
\usepackage{colortbl}  
\newcommand{\etal}{{\em et al.\,}}       
\newcommand{\eg}{{\em e.g.}}           
\newcommand{\ie}{{\em i.e.}}           
  
\definecolor{tblblue}{RGB}{129,184,223}
\definecolor{tblred}{RGB}{254,129,125}

\AtBeginDocument{%
  \providecommand\BibTeX{{%
    \normalfont B\kern-0.5em{\scshape i\kern-0.25em b}\kern-0.8em\TeX}}}

\setcopyright{acmlicensed}
\copyrightyear{2024}
\acmYear{2024}
\acmDOI{}

\acmSubmissionID{2024-0694}
\begin{document}

\title{A SAM-guided Two-stream Lightweight Model for Anomaly Detection}

\author{Chenghao Li}
\email{lichenghao@seu.edu.cn}
\orcid{0009-0003-6130-1210}

\author{Lei Qi}
\authornotemark[1]
\email{qilei@seu.edu.cn}

\author{Xin Geng}
\email{xgeng@seu.edu.cn}
\affiliation{%
  \institution{School of Computer Science and Engineering (Southeast University), and Key Laboratory of New
Generation Artiicial Intelligence Technology and Its Interdisciplinary Applications (Southeast University),
Ministry of Education}
  \city{Nanjing}
  \state{Jiangsu}
  \country{China}
  \postcode{210019}
}
\thanks{Corresponding author: Lei Qi}

\thanks{The work is supported by NSFC Program (Grants No. 62206052, 62125602, 62076063), China Postdoctoral Science Foundation (Grants No. 2024M750424), Supported by the Postdoctoral Fellowship Program of CPSF (Grant No. GZC20240252), and Jiangsu Funding Program for Excellent Postdoctoral Talent (Grant No. 2024ZB242).}


\renewcommand{\shortauthors}{Li and Qi, et al.}

\begin{abstract}
In industrial anomaly detection, model efficiency and mobile-friendliness become the primary concerns in real-world applications. Simultaneously, the impressive generalization capabilities of Segment Anything (SAM) have garnered broad academic attention, making it an ideal choice for localizing unseen anomalies and diverse real-world patterns. In this paper, considering these two critical factors, we propose a SAM-guided Two-stream Lightweight Model for unsupervised anomaly detection (STLM) that not only aligns with the two practical application requirements but also harnesses the robust generalization capabilities of SAM. We employ two lightweight image encoders, \ie, our two-stream lightweight module, guided by SAM's knowledge. To be specific, one stream is trained to generate discriminative and general feature representations in both normal and anomalous regions, while the other stream reconstructs the same images without anomalies, which effectively enhances the differentiation of two-stream representations when facing anomalous regions. Furthermore, we employ a shared mask decoder and a feature aggregation module to generate anomaly maps. Our experiments conducted on MVTec AD benchmark show that STLM, with about 16M parameters and achieving an inference time in 20ms, competes effectively with state-of-the-art methods in terms of performance, 98.26\% on pixel-level AUC and 94.92\% on PRO. We further experiment on more difficult datasets, \eg, VisA and DAGM, to demonstrate the effectiveness and generalizability of STLM. Codes are available online at \url{https://github.com/Qi5Lei/STLM}.
\end{abstract}



\received{03 July 2024}
\received[revised]{13 October 2024}
\received[accepted]{18 November 2024}

\maketitle

\section{Introduction}
Image anomaly detection and localization tasks have attracted great attention in various domains, \eg, industrial quality control~\cite{bergmann2019mvtec,roth2022towards}, medical diagnoses~\cite{Medical2021,SHI2023Combating}, and video surveillance~\cite{Exploring2019,Surveillance2017}. These tasks aim to discriminate both abnormal images and anomalous pixels in images according to previously seen normal or pseudo-anomalous samples during training. However, anomaly detection and localization are especially hard, as anomalous images occur rarely and anomalies can vary from subtle changes to large defects such as broken parts. Creating a dataset that includes sufficient anomalous samples with all possible anomaly types for training is a formidable challenge. Therefore, most methods have turned to tackling AD tasks with unsupervised models~\cite{tien2023revisiting, deng2022anomaly, Rethinking2020}, by only relying on normal samples, which are of great significance in practice. 

\begin{wrapfigure}{R}{0.57\textwidth} 
\centering
\includegraphics[width=0.57\columnwidth]{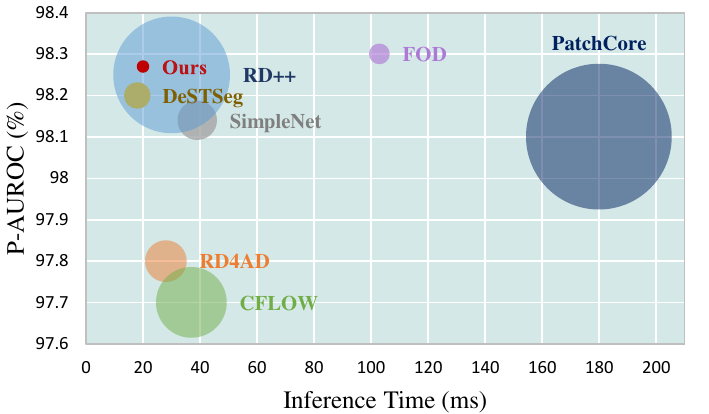}
\caption{Comparisons of different anomaly detection methods in terms of pixel-level AUROC (vertical axis), inference time (horizontal axis), and the ratios of parameter numbers (circle radius). Our STLM achieves competitive pixel-level AUROC for anomaly detection while being 8× faster than \textcolor[RGB]{0,31,95}{PatchCore}, \textbf{4× faster} than \textcolor[RGB]{174,121,213}{FOD} which achieves the highest pixel-level AUROC, 1× faster than \textcolor[RGB]{148,148,148}{SimpleNet}, and 0.5× faster than \textcolor[RGB]{31,56,99}{\textbf{RD++ (154.87M)}}. In addition, STLM requires only \textbf{16.56M} of parameters for inference, making it one of the most efficient methods.}
\label{fig1}
\end{wrapfigure}

To achieve this, the utilization of a memory bank is proposed, in which a core set stores features extracted from the pre-trained backbone, to calculate a patch-level distance between the core set and the sample~\cite{yao2023focus,zhang2023prototypical}. However, these methods, which involve creating memory banks, come at the cost of increased computational complexity and large memory space. 

Recent efforts have been directed towards proposing effective anomaly detection methods, with particular emphasis on reconstruction-based methods~\cite{Robust2023}, which are favored for their simplicity and interpretability. These methods typically employ autoencoders~\cite{bergmann2019improving,Yi2023Repre,zhou2017},variational autoencoders~\cite{Liu2020}, or generative adversarial networks~\cite{goodfellow2014generative}. Based on the assumption that the reconstruction error of normal regions is lower than that of abnormal ones, they compare the test images with their reconstructed counterparts. However, recent studies have revealed that deep models can generalize so effectively that they accurately restore anomalous regions~\cite{zavrtanik2021reconstruction}, undermining the effectiveness of anomaly detection tasks. 

To address this challenge, memory modules have been integrated into reconstruction-based methods~\cite{gong2019memorizing}. These modules store representations of normal images, and the representations of test images are used as queries to retrieve the most relevant memory items for reconstruction. However, they suffer from high memory requirements and search times.

The knowledge distillation (KD)~\cite{hinton2015distilling} based framework has shown its effectiveness in anomaly detection and localization~\cite{tien2023revisiting,roth2022towards,salehi2021multiresolution}. 
For example, Salehi \etal~\cite{salehi2021multiresolution} established an S-T network pair where knowledge is transferred from the teacher to the student. The underlying hypothesis is that the student network sees only the normal samples in the training stages, leading to generating out-of-distribution representations with anomalous queries during inference. However, the statement does not always hold true due to architectural similarities and shared data flow~\cite{deng2022anomaly}. To overcome this limitation, DeSTSeg~\cite{zhang2023destseg} introduces a denoising student network to generate distinct feature representations from those of the teacher when handling anomalous inputs.

Recently, foundation models, \eg, Segment Anything (SAM)~\cite{kirillov2023segment} have demonstrated great zero-shot abilities through the retrieval of prior knowledge. SAM is trained on millions of annotated images, enabling it to generate high-quality segmentation results for previously unseen images. Nevertheless, foundation models often have limitations within certain domains~\cite{ji2023segment} and SAM contains 615M parameters against the mobile-friendly requirement.

Considering the generality problem of unseen anomalies and the real-world application requirements, we propose a SAM-guided Two-stream Lightweight Model for unsupervised anomaly detection (\textbf{STLM}). It takes advantage of the robust generalization capabilities of foundation models and aligns with the mobile-friendly requirements. As illustrated in Figure~\ref{fig2}, STLM consists of a fixed SAM teacher, a trainable two-stream lightweight model (\textbf{TLM}), and a feature aggregation (\textbf{FA}) module. We start with Pseudo Anomaly Generation prepossess on the normal training images to balance the number of normal and anomalous images.

After processing the data, the TLM is introduced. First, instead of directly utilizing the pre-trained teacher network during inference in KD, we suggest distilling the comprehensive knowledge from the fixed SAM to a student stream as the lightweight ``teacher'', called the plain student stream, which is more precise and generalized for representing features related to anomaly detection (AD) task. Second, inspired by~\cite{zhang2023destseg}, we incorporate the training of a denoising stream, also distilled from the SAM. The denoising student stream takes a pseudo-anomalous image as input, whereas the teacher SAM operates on the original, clean image. Our method effectively enhances the differentiation of two-stream features when dealing with anomalous regions. The impressive capabilities of SAM alleviate the generality problem typically associated with being opaque to genuine anomalies during training and various normal patterns in practical scenarios. In our method, we employ the encoder of the MobileSAM~\cite{zhang2023faster} as the backbone for our TLM's encoder. Besides, we design a shared mask decoder and a feature aggregation module to generate anomaly maps.
 
We evaluate our method on the MVTec AD benchmark~\cite{bergmann2019mvtec}, which is specifically designed for anomaly detection and localization. Extensive experimental results show that our method is competitive with state-of-the-art while tackling the task of image-level and pixel-level anomaly detection. With remarkable parameter efficiency and swift inference speed, our STLM achieves competitive results on MVTec, VisA and DAGM. In order to validate the effectiveness of our proposed components, we conduct comprehensive ablation studies. 
The contributions of the proposed method are highlighted as follows:

\begin{enumerate}
\item[$\bullet$] We propose a SAM-guided Two-stream Lightweight Model for unsupervised anomaly detection that not only conforms to the model efficiency and mobile-friendliness demands of practical industrial applications but also takes advantage of the robust generalization capabilities of SAM for effectively exploring unseen anomalies and diverse normal patterns.
\item[$\bullet$] We conduct extensive experiments on MVTec, VisA and DAGM. Results show that our method with about 16M parameters is competitive with state-of-the-art methods on detection and localization, underscoring its robust generalization capabilities. Notably, taking both performance and model parameter size into consideration, our method is a promising solution for practical applications.
\end{enumerate}

\section{Related Work}
In this section, we review some related works to our work, including Deep Learning Methods for Anomaly Detection and Localization, Vision Foundation Models and Lightweight Models.

\subsection{Deep Learning Methods for Anomaly Detection and Localization}
 Prior works of anomaly detection and localization have explored various methodologies. The method of \textbf{image reconstruction} is widely adopted, which posits that accurately reconstructing anomalous regions can be challenging due to their absence in the training samples. Generative models such as autoencoders~\cite{bergmann2019improving,AE2021}, variational autoencoders~\cite{VAE2023}, and generative adversarial networks~\cite{goodfellow2014generative,GAN2019} are also utilized to reconstruct normal images from anomalous ones. Nonetheless, these methods face certain limitations, especially when reconstructing complex textures and patterns. Later methods use deep models to enhance the quality of reconstructing images~\cite{zavrtanik2021draem,zavrtanik2021reconstruction}. 

Recently, the utilization of a \textbf{memory bank}, in which a core set stores features extracted from the pre-trained backbone, calculates a patch-level distance between the core set and the sample to detect anomalies~\cite{yao2023focus}. 
The studies in~\cite{zhang2023prototypical} reveal a method that focuses on learning feature residuals of varying scales and sizes between anomalous and normal patterns, which prove beneficial in accurately reconstructing the segmentation maps of anomalous regions.
However, these methods, which involve creating memory banks, come at the cost of increased computational complexity.

\textbf{Knowledge distillation}~\cite{hinton2015distilling} relies on a pre-trained teacher network and a trainable student network. Since the student network is trained on anomaly-free samples, it is expected to have feature representations that differ from those of the teacher network~\cite{bergmann2020uninformed,salehi2021multiresolution,wang2021student}. 
For instance, to capture anomalies at multiple scales, multi-resolution knowledge distillation~\cite{salehi2021multiresolution} is used to distinguish unusual features on multi-level features. 
Prior studies aim to enhance the similarity of the features when processing normal images, whereas DeSTSeg tries to separate their representations specifically when dealing with anomalous regions. However, the teacher network is hard to accurately represent unseen normal textures and patterns with some variations and the denoising student network has generality limitations of being opaque to genuine anomalies. 

\subsection{Vision Foundation Models}
Foundation models~\cite{bommasani2021opportunities,brown2020language} initially emerged, that large language models such as the GPT series~\cite{li2023evaluation} showcased impressive zero-shot generalization~\cite{lin2022ensemble,zheng2022faster,WeiGAO2023} to unseen tasks and data. This led to the development of prompt-based learning works~\cite{houlsby2019parameter, Hu2021LoRALA} aimed at helping these pre-trained models generalize to downstream tasks instead of fine-tuning their internal parameters~\cite{hinterstoisser2018pre} for better transfer learning.

For vision-based foundation models~\cite{ji2023segment,wang2023seggpt,zou2024segment}, prompt engineering~\cite{zhou2023zegclip}— freezing the pre-trained model— was first explored in vision-language models like CLIP. \textbf{CLIP}~\cite{radford2021learning}, which was trained on 400 million image-text pairs, can represent visual content through textual prompts. It was demonstrated strong zero-shot performance on major benchmark datasets. However, CLIP's performance diminishes when dealing with images not included in its pre-training datasets, highlighting the impact of training data size and distribution on downstream task performance. Although CLIP and CLIP-style models facilitate vision tasks by linking images with descriptive text to capture scene-level semantics, they struggle with identifying instance-level semantics within scenes.

\textbf{DINO}~\cite{caron2021emerging} utilizes a self-supervised student-teacher framework based on vision transformers~\cite{dosovitskiy2021an}. It learns robust and high-quality features for downstream tasks due to employing a multi-view strategy. Recently, further advances have been shown by \textbf{SAM}~\cite{kirillov2023segment} and \textbf{DINOv2}~\cite{oquab2023dinov2}. DINOv2 integrates ideas from DINO with patch-level reconstruction methods~\cite{zhou2022image}. SAM, trained on millions of annotated images, generated high-quality segmentation results for previously unseen images. Their capabilities have been evaluated across numerous tasks, showcasing impressive zero-shot performance that is often competitive with or even superior to prior fully supervised methods. Additionally, we aim to leverage SAM's exceptional segmentation strength for our AD tasks. 

However, foundation models often exhibit limitations in specific domains, including medical imaging and anomaly detection~\cite{ji2023segment}. Since they are mostly transformer-based frameworks, they also have issues with computation complexity and inference latency at inference time~\cite{pope2023efficiently}. And they are mostly trained in natural images, they tend to prioritize foreground objects and may struggle to accurately segment small or irregular objects. Therefore, our work seeks to explore how to effectively harness the extensive knowledge these off-the-shelf models provide.

\subsection{Lightweight Models}
While deep learning offers significant advantages for feature extraction, traditional neural networks often demand substantial computational power and memory resources during training. In recent years, there has been notable progress in developing lightweight models, particularly for deployment on resource-constrained devices. To address these challenges, techniques like \textbf{model pruning}~\cite{Wang2024DCFP,Liu2019MetaPruning}, \textbf{quantization}~\cite{Yao2021HAWQ,Liu2021Quantization}, and \textbf{knowledge distillation}~\cite{hinton2015distilling,salehi2021multiresolution} have been explored to reduce model size and computational complexity while maintaining performance. CNN-based lightweight architectures such as MobileNet~\cite{howard2017mobilenets}, ShuffleNet~\cite{zhang2018shufflenet}, and CSPNeXt~\cite{Chen2024CSPNeXt} have introduced innovations like depthwise separable convolutions and group convolutions, significantly reducing inference time and memory requirements.

Despite these advancements, challenges persist particularly in unsupervised tasks like anomaly detection, where accuracy, efficiency and generalization are crucial. State-of-the-art CNN-based methods, such as SimpleNet~\cite{liu2023simplenet} and DeSTSeg, achieve satisfactory performance but fall short in generalization across multiple datasets. Transformers, with their large amount of parameters, offer high interpretability through self-attention mechanisms~\cite{Cao2024Self}. It also captures global dependencies, in contrast to CNNs which tend to emphasize local features. These properties help Transformers to a more suitable architecture for unsupervised tasks, particularly in training generalized vision models on large-scale datasets, such as SAM and DINOv2. Recent works, proposing efficient Transformers, have further enhanced their applicability in resource-constrained environments.

\begin{figure}%
	\centering
\includegraphics[width=1\columnwidth]{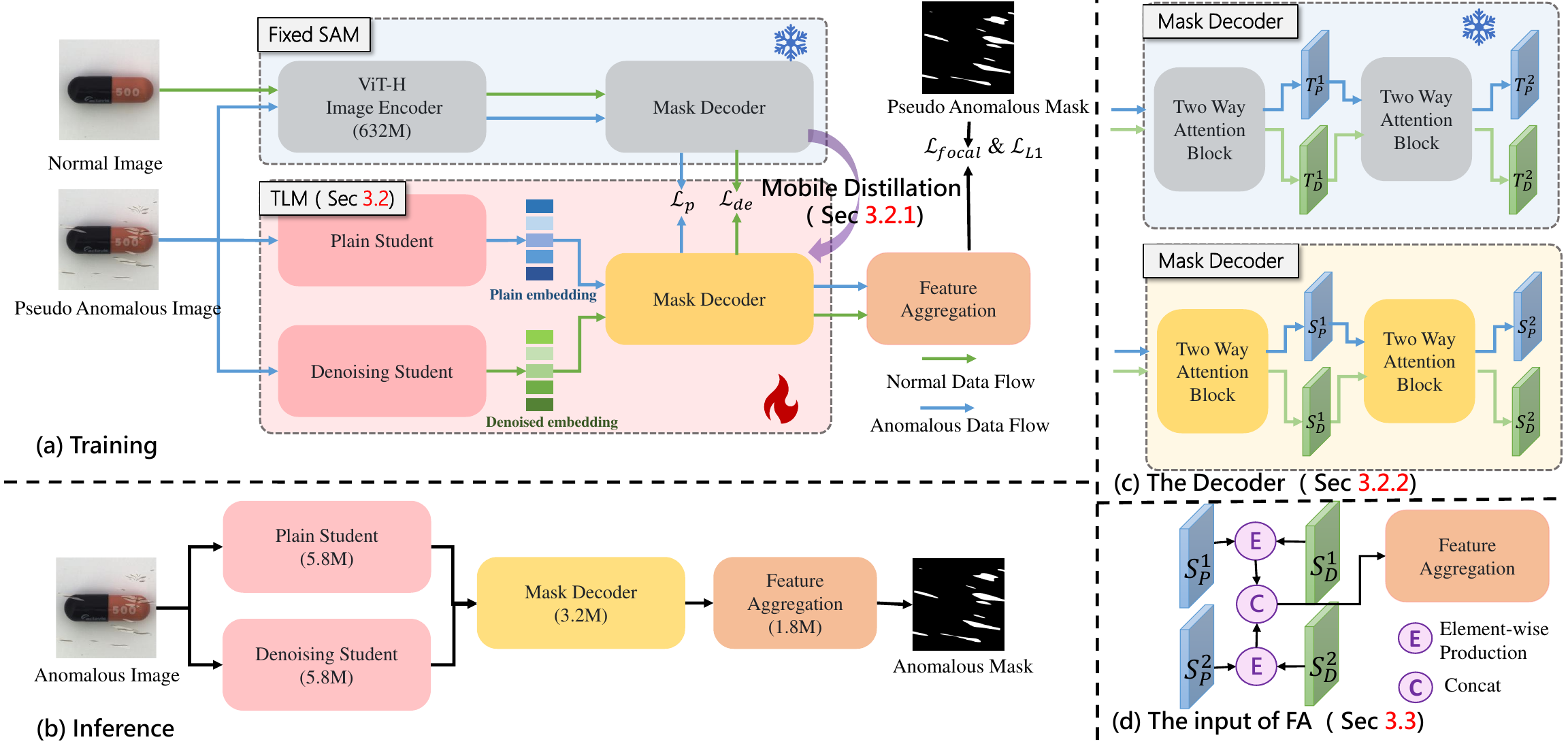}
\caption{Overview of STLM. Pseudo anomalies are introduced into normal training images with predefined probabilities and are used exclusively during \textbf{training}. The trainable two-stream lightweight model (TLM) distills different information from a fixed SAM teacher. One stream is trained to generate discriminative and generalized feature representations in both normal and anomalous regions, while the other to match its feature of the same images without corruption, which enhances the differentiation of their representations when addressing anomalous regions. The element-wise product of the TLM is employed to train the feature aggregation module with the generated binary anomaly mask. For \textbf{inference}, anomaly maps are generated only using the TLM and FA module.}
	\label{fig2}
\end{figure}

\section{Method}
Together with the pseudo anomalies introduced into normal training images with predefined probabilities (in Section~\ref{sec: pseudo anomaly}) to maintain balanced data distribution, we propose the SAM-guided Two-stream Lightweight Model for unsupervised anomaly detection (STLM) to effectively generate an anomaly map for anomaly detection and localization. As illustrated in Figure~\ref{fig2}, the two-stream lightweight model (TLM, in Section~\ref{sec: SKLST}), consisting of a plain student stream and a denoising student stream, distills different information from a fixed SAM~\cite{kirillov2023segment} during training. Consequently, a feature aggregation module (FA, in Section~\ref{sec: FA}) is trained to fuse anomaly features with the aid of additional supervision signals.
For inference (in Section~\ref{sec: inference}), anomaly maps are generated solely using the TLM and the FA module. 

\subsection{Pseudo Anomaly Generation}
\label{sec: pseudo anomaly}

To address the issue of data imbalance, the training of our model relies on diverse types of pseudo-anomalous images, generated using the methodology introduced in~\cite{zavrtanik2021draem}.
A noise image is generated using a Perlin noise generator, and then Binarization is applied to the Random Perlin noise to acquire an anomaly mask $M$. The proposed pseudo anomaly image $I_{a}$ can be defined as follows:
\begin{equation}\label{eqn-1} 
  I_{a} = \bar{M} \odot N + (1 - \beta)(M \odot A) + \beta (M \odot N ) ,
\end{equation}
where $N$ is the normal sample, $A$ is the external data source an arbitrary image from, $\bar{M}$ is the inverse of $M$, $\odot$ means the element-wise multiplication operation. $\beta$ is the opacity parameter for a better combination of abnormal and normal regions. In practice, we use a probability of 0.5 to decide if this generation method is activated or not, with the ablation study shown in Figure~\ref{fig4_c}. Notably, the generation process is conducted in real-time training. 

\subsection{Two-Stream Lightweight Model}
\label{sec: SKLST}
In traditional KD, the teacher network typically employs a large and deep expert network with extensive capabilities, while the student network adopts a similar neural network structure to that of the teacher~\cite{bergmann2019improving}, which cannot learn about anomalous regions, leading to discrepancies with the teacher network during testing. However, extensive experiments have shown that this structure can cause collapsing (constantly generating the same normal image). Therefore, the reconstructed image is used for anomaly discrimination, while the teacher generator acts as a regularizer to prevent this issue. Reconstructing anomalous regions as normal reduces the student network's ability to “discriminate” anomalous areas, thereby increasing the recognition difference between the teacher and student networks. Despite~\cite{zhang2023destseg} introducing a denoising student network to address this issue, it still fails to consider that the teacher network struggles to accurately represent unseen normal textures and patterns with some variations compared to the training set. The denoising student network has generality limitations of being opaque to genuine anomalies while training. 


\subsubsection{\textbf{Mobile Distillation}}
SAM~\cite{kirillov2023segment} has attracted notable academic interest owing to its remarkable zero-shot capabilities and versatility across various vision applications. In our paper, we aim to harness the robust generalization abilities of SAM and render SAM more mobile-friendly. 

First, the TLM is a variation of the commonly used Student-Teacher(ST) distillation network in Anomaly Detection. In a conventional ST network, the student network, trained only on normal images, cannot learn about anomalous regions, leading to discrepancies with the teacher network during testing. However, extensive experiments have shown that this structure can cause collapse (constantly generating the same normal image). Therefore, the reconstructed image is used for anomaly discrimination, while the teacher generator acts as a regularizer to prevent this issue. Reconstructing anomalous regions as normal reduces the student network's ability to “discriminate” anomalous areas, thereby increasing the recognition difference between the teacher and student networks. This approach has also been studied in~\cite{bergmann2020uninformed,zhang2023destseg}. Specifically, the plain student stream is trained to produce discriminative and general feature representations in both normal and anomalous regions as a superior alternative to the teacher network, demonstrated in Section~\ref{sec: ablation}. Similarly guided by SAM's knowledge, the denoising student stream aligns its feature representations with those of the same images without any corruption, which effectively enhances the differentiation of the representations from two student streams when dealing with anomalous regions.

Notably, based on the results from Section~\ref{sec: ablation}, where one-stage training outperforms two-stage training, we explain why it performs better: the FA and TLM modules being trained simultaneously. On the one hand, the trained plain stream, compared to the frozen SAM teacher network, exhibits greater adaptability to other parts of TLM, which generates better intermediate results feeding to the FA module. On the other hand, segmentation knowledge related to anomaly detection images is feedback through the gradients of the FA module, back into both streams to better recognize normal regions. Only the denoising student is improved without distillation training for SAM, resulting in differences in normal regions and affecting the experimental results. 

Second, the principal aim of our paper is to develop a model that can deliver satisfactory performance while significantly reducing the number of parameters and inference time. SAM has demonstrated its ability to function on resource-constrained devices, primarily thanks to its lightweight mask decoder.  However, the default image encoder in the original SAM relies on ViT-H, which boasts over 600M parameters and is considered heavyweight. Considering the 3.2M decoder lightweight enough, We transition the default image encoder to ViT-Tiny~\cite{wu2022tinyvit} while keeping the original architecture of the mask decoder.

Consequently, SAM serves as the teacher network for the TLM. The output feature maps are extracted from the two-layer decoder of SAM, shown in Figure~\ref{fig2} (c). With no need for segmentation outcomes, we only use the Two-Way-Transformer blocks of the mask decoder, and introduce our FA module. The output features from the pre-trained SAM with normal images as input are denoted as $T_{D}^{1}$ and $T_{D}^{2}$, respectively. Similarly, those originating from pseudo-anomalous images are labeled $T_{P}^{1}$ and $T_{P}^{2}$. TLM adopts the image encoder of mobileSAM~\cite{zhang2023faster}, \ie, ViT-Tiny~\cite{wu2022tinyvit}, and their output features are denoted as $S_{P}^{1}$, $S_{P}^{2}$, $S_{D}^{1}$, and $S_{D}^{2}$, respectively.

To distill knowledge from SAM to the plain student stream, we minimize the cosine distance between features from $T_{P}^{k}$ and $S_{P}^{k}$, where $k$ = 1,2. Additionally, we minimize the cosine distance between features from $T_{D}^{k}$ and $S_{D}^{k}$, for $k$ = 1,2, to supervise the denoising student stream in reconstructing normal features. The cosine similarity can be computed through Equation~\eqref{eqn-2} and the loss function for optimizing the network is formulated as Equation~\eqref{eqn-3} and Equation~\eqref{eqn-4}.
\begin{equation}\label{eqn-2} 
  X_{k}(i,j) = \frac{T_{k}(i,j) \odot S_{k}(i,j)}{||T_{k}(i,j)||_{2}||S_{k}(i,j)||_{2}} ,
\end{equation}
\begin{equation}\label{eqn-3} 
  \mathcal{L}_{p} = \sum_{k=1}^{2}\left\{\frac{1}{H_{k}W_{k}}\sum_{i,j=1}^{H_{k},W_{k}}(1-X_{k}^{P}(i,j))\right\} ,
\end{equation}
\begin{equation}\label{eqn-4} 
  \mathcal{L}_{de} = \sum_{k=1}^{2}\left\{\frac{1}{H_{k}W_{k}}\sum_{i,j=1}^{H_{k},W_{k}}(1-X_{k}^{D}(i,j))\right\} ,
\end{equation}
where $k$ is the number of feature layers used in training. 
$i$ and $j$ stand for the spatial coordinate on the feature map.In particular, $i = 1\ldots H_{k}$ and $j = 1\ldots W_{k}$. $H$ and $W$ denote the height and width of $k$-th feature map.
The cosine similarity between features from SAM and the plain stream denotes $X_{k}^{P}(i,j)$, while that between features from SAM and the denoising stream denotes $X_{k}^{D}(i,j)$.


\subsubsection{\textbf{The Decoder}} 
Inspired by~\cite{zhang2023faster}, the image embeddings generated by the student encoder can closely approximate that of the original teacher encoder. This observation leads to the conclusion that the utilization of distinct decoders for the TLM may not be imperative. Instead, a shared two-layer decoder, is an efficient alternative. We observe that the shared decoder, while economical in terms of parameters and computation time, achieves a satisfactory result with a slight decrease compared to employing two separated decoders, shown in Figure~\ref{fig4_a}.

\subsubsection{\textbf{Remark}} 
The difference between our STLM and DeSTSeg~\cite{zhang2023destseg} lies in 1) our method learns not only a denoising student but also a plain student in our TLM. The plain student stream can learn generalized knowledge related to the anomaly detection task, which enables it to effectively represent normal features and even previously unseen patterns in the training set, ensuring differences between features of two streams are well-captured. 2) The high-generalization-capable SAM, as the teacher network of our TLM, also helps the denoising student to generate high-quality reconstructed results for genuine anomalies. Besides, it is worth noting that our method exhibits superior performance and a more compact model compared to DeSTSeg.

\subsection{Feature Aggregation Module}
\label{sec: FA}
Previous studies~\cite{salehi2021multiresolution,wang2021student} have noted that sub-optimal results can occur when distinctions among features at different levels are not uniformly precise.  Following an extensive series of experiments, it has become evident that the inclusion of a segmentation network, guiding feature fusion through additional supervision signals, improves performances. The feature aggregation (FA) module is composed of two residual blocks and an atrous spatial pyramid pooling (ASPP) module. As a primary consideration, we have refrained from constraining the weights of the TLM when training the segmentation network. 

Notably, our findings indicate that simultaneous training of these components yields enhanced results. Furthermore, it has been observed that reducing the depth of the FA module does not significantly impact the performance as indicated in Figure~\ref{fig4_b}. As a result, an effort has been made to make this module more lightweight by reducing the channels from 256 to 128 and adjusting the dilation rate to [1, 1, 3], following a similar method in~\cite{zhang2023cross}. 
It mitigates memory costs associated with both training and inference, which is a factor of paramount importance in practical implementations. 

When training, synthetic anomalous images are employed as inputs for the TLM, with the corresponding binary anomaly mask serving as the ground truth. The decoder generates features that have the same shape as the ground truth mask $M$. The similarities are calculated using Equation~\eqref{eqn-2} and then concatenated as $\hat{X}$, which is subsequently fed into the FA module, shown in Figure~\ref{fig2} (d). Inspired by~\cite{zavrtanik2021draem}, a focal loss and an L1 loss are applied to increase the robustness toward accurate segmentation of challenging examples and reduce over-sensitivity to outliers, respectively.
The FA module outputs an anomaly score map $M_{ij}^{o}$, which is of the same shape as the ground truth mask $M$.
\begin{equation}\label{eqn-5} 
  p_{ij} = M_{ij}M_{ij}^{o} + (1-M_{ij})(1-M_{ij}^{o}) ,
\end{equation}
\begin{equation}\label{eqn-6} 
  \mathcal{L}_{focal} = -\frac{1}{HW}\sum_{i,j=1}^{H,W}(1-p_{ij})^\gamma \log(p_{ij}) ,
\end{equation}
\begin{equation}\label{eqn-7} 
  \mathcal{L}_{l1} = -\frac{1}{HW}\sum_{i,j=1}^{H,W}|M_{ij}-M_{ij}^{o}| ,
\end{equation}
where the focus ($\gamma$) is set as 4, following the set in~\cite{zhang2023prototypical,zhang2023destseg}. 

\subsection{Training and Inference}
\label{sec: inference}
The training stage is described in Figure~\ref{fig2} (a), and the loss functions used are proposed in Section~\ref{sec: SKLST} and Section~\ref{sec: FA}.  
\begin{equation}\label{eqn-8} 
  \mathcal{L}_{total} = \mathcal{L}_{p}+\mathcal{L}_{de}+\mathcal{L}_{focal}+\mathcal{L}_{l1} ,
\end{equation}
where each loss function carries equal weight.

For inference, the fixed SAM is discarded and the procedure is presented in Figure~\ref{fig2} (b). For pixel $ij$, the pixel-level anomaly segmentation map $M_{ij}$ is provided by the end of the network. It is anticipated that the output will have higher values for pixels that are anomalous pixels. For the computation of the image-level anomaly score, we take the average of the top-$K$ anomalous pixel values from the anomaly score map, following~\cite{zhang2023prototypical,zhang2023destseg}.

\begin{figure}%
	\centering
\includegraphics[width=1\columnwidth]{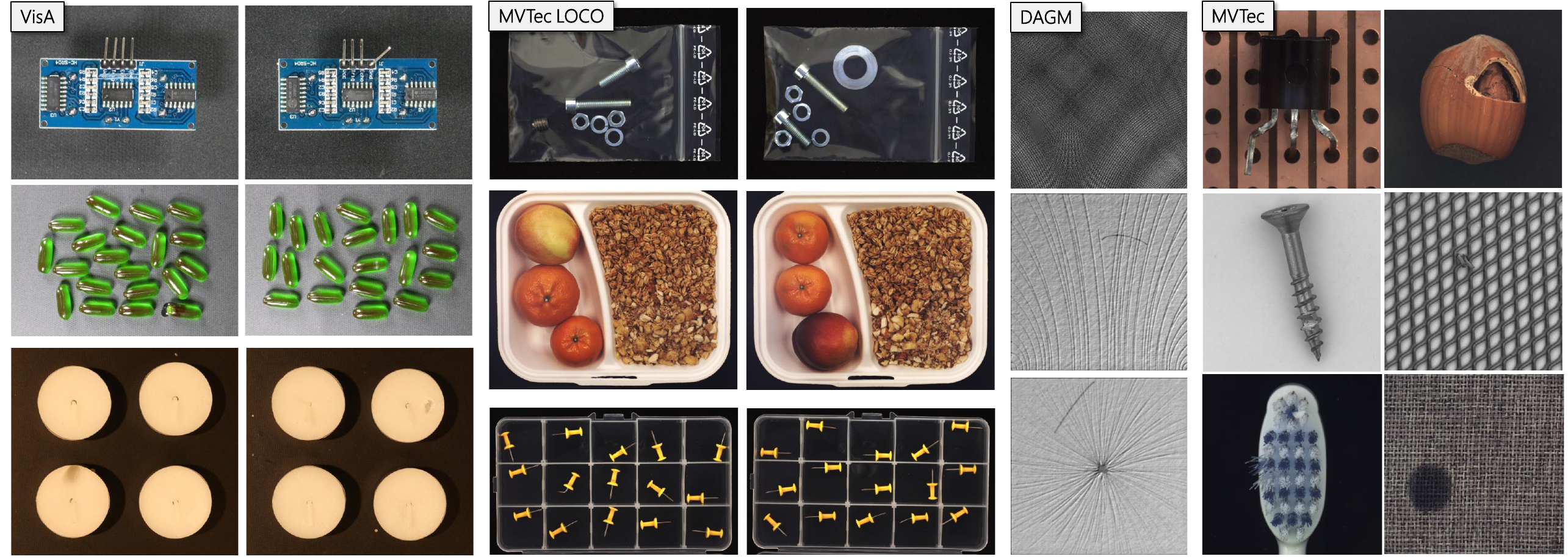}
\caption{Visualization examples of VisA~\cite{zou2022spot}, MVTec LOCO~\cite{Loco2022}, DAGM~\cite{zavrtanik2021draem} and MVTec~\cite{bergmann2019mvtec}.}
	\label{fig13}
\end{figure}

\section{Experiments}
\label{sec:experiments}
\subsection{Experimental Details}
\label{sec: details}
\subsubsection{\textbf{Datasets}}
\label{sec: datasets}
We validate the effectiveness of our method using the MVTec AD~\cite{bergmann2019mvtec} dataset, a renowned benchmark in the field of anomaly detection and localization. MVTec AD comprises 5 texture and 10 object categories, each providing hundreds of normal images for training, along with a diverse set of both anomalous and normal images for evaluation. It also provides pixel-level ground truths for defective test images.

Additionally, we extend our experimentation on the VisA~\cite{zou2022spot}, MVTec LOCO~\cite{Loco2022} and DAGM~\cite{wieler2007weakly} dataset to showcase the generalization capabilities of STLM facing more complex datasets with, examples in Figure~\ref{fig13}. The VisA Dataset comprises 10,821 high-resolution color images (9,621 normal and 1,200 anomalous samples) encompassing 12 objects across 3 domains, establishing it as the most extensive anomaly detection dataset in the industrial domain to date. The MVTec LOCO dataset includes both structural and logical anomalies. It contains 3644 images from five different categories inspired by real-world industrial inspection scenarios. Structural anomalies appear as scratches, dents, or contaminations in the manufactured products. Logical anomalies violate underlying constraints, e.g., a permissible object being present in an invalid location or a required object not being present at all. Notably, for the convenience of comparison, we only select structural anomalies. DAGM contains 10 textured objects with small abnormal regions that bear a strong visual resemblance to the background. For each class, we first move all anomalous samples from the original training set to the original test set and then move all normal samples from the test set to the training set. Furthermore, we randomly select 30 normal samples from the training set, designating them as "good" samples in the test set. 

\subsubsection{\textbf{Evaluation Metrics}}
\label{sec: metrics}
In line with previous research, we employ AUC (Area Under the ROC Curve) to evaluate image-level and pixel-level anomaly detection. However, anomalous regions typically only occupy a tiny fraction of the entire image. Consequently, Pixel-AUROC may not accurately reflect the localization accuracy, as the false positive rate is primarily influenced by the vast number of anomalous-free pixels~\cite{tao2022deep}. To offer a more comprehensive measure of localization performance, we introduce the Per Region Overlap (PRO)~\cite{bergmann2020uninformed}, which assigns equal weight to anomaly regions of varying sizes. Pixel-level Average Precision (AP) is also introduced for comprehensive measuring. The PRO score treats anomaly regions of varying sizes equally, whereas AP is more suitable for highly imbalanced classes, particularly in the context of industrial anomaly localization where accuracy plays a critical role. Our comparison SOTA methods include DRAEM~\cite{zavrtanik2021draem}, CFLOW~\cite{gudovskiy2022cflow}, CFA~\cite{lee2022cfa}, PatchCore~\cite{roth2022towards}, RD4AD~\cite{deng2022anomaly}, SimpleNet~\cite{liu2023simplenet}, DeSTSeg~\cite{zhang2023destseg}, RD++~\cite{tien2023revisiting}, FastFlow~\cite{yu2021fastflow}, and FOD~\cite{yao2023focus}.

\subsubsection{\textbf{Implementation Details}}
\label{sec: detail}
Our initialized two-stream encoders of TLM are the ViT-Tiny~\cite{wu2022tinyvit} with MoblieSAM weight, while SAM serves as the teacher network. All images in the two datasets are resized to 1024 × 1024. We employ the Adam Optimizer, setting the learning rate at 0.0005 for the TLM. For the FA module, we choose the Stochastic Gradient Descent Optimizer, using the same configuration as presented in~\cite{zhang2023destseg}. Each loss function in our paper carries equal weight. We conduct training for 200 epochs with a batch size of 2 and compute the average of the top 100 anomalous pixels as the image-level anomaly score. Notably, we use the augmentation method, \ie, pseudo anomaly generation strategy, that most state-of-the-art methods use for fair competition. Prototypical~\cite{zhang2023prototypical} presents more complex anomaly generation strategies to performe better.

\begin{table}
    \caption{Anomaly detection and location results in terms of AUROC ($\%$) at image-level and pixel-level on the MVTec dataset~\cite{bergmann2019mvtec}}
    \begin{tabular}{lcccccccccccc}
    \toprule
         \; \; Method & DRAEM~\cite{zavrtanik2021draem} & CFLOW~\cite{gudovskiy2022cflow} & PatchCore~\cite{roth2022towards} & RD4AD~\cite{deng2022anomaly} & Ours \\
    \midrule
    Carpet & 96.90/97.50 & 97.60/99.20 & 99.10/99.00 & 98.70/98.90 &  99.48/\textbf{99.91} \\
    Grid  & 99.90/\textbf{99.70} & 98.10/98.90 & 97.30/98.70 & \textbf{100}/98.30 & 95.57/95.40 \\
    Leather & \textbf{100}/99.00 & 99.90/\textbf{99.70} & \textbf{100}/99.30 & \textbf{100}/99.40 &  \textbf{100}/99.11 \\
    Tile  & \textbf{100}/\underline{99.20} & 97.10/96.20 & 99.30/95.80 & 99.70/95.70 & \textbf{100}/\textbf{99.58} \\
    Wood  & 99.50/95.50 & 98.70/86.00 & 99.60/95.10 & 99.50/95.80 & \textbf{100}/\underline{97.99} \\
    Bottle & 98.00/\underline{99.10} & 99.90/97.20 & \textbf{100}/98.60 & \textbf{100}/98.80 &  \textbf{100}/97.77 \\
    Cable & 90.90/95.20 & 97.60/97.80 & \textbf{99.90}/\textbf{98.50} & 96.10/97.20 & 98.98/96.95 \\
    Capsule & 91.30/88.10 & 97.00/\textbf{99.10} & 98.00/99.00 & 96.10/98.70 & 98.69/98.48 \\
    Hazelnut & \textbf{100}/\textbf{99.70} & \textbf{100}/98.80 & \textbf{100}/98.70 & \textbf{100}/99.00 &  \textbf{100}/ \underline{99.69} \\
    Metal\_nut & \textbf{100}/\textbf{99.60} & 98.50/98.60 & 99.90/98.30 & \textbf{100}/97.30 & \textbf{100}/\underline{99.45} \\
    Pill  & 97.10/97.30 & 96.20/\underline{98.90} & 97.50/97.60 & \underline{98.70}/98.10 & 98.20/98.61 \\
    Screw & 98.70/99.30 & 93.10/98.90 & 98.20/99.50 & 97.80/\textbf{99.70} & \textbf{99.47}/97.56 \\
    Toothbrush & \textbf{100}/97.30 & 98.80/99.00 & \textbf{100}/98.60 & \textbf{100}/99.10 & 99.06/\textbf{99.61} \\
    Transistor & 91.70/85.20 & 92.90/\textbf{98.20}& 99.90/\underline{96.50} & 95.50/92.30 & 97.58/94.81 \\
    Zipper & \textbf{100}/\underline{99.10} & 97.10/\underline{99.10} & 99.50/98.90 & 97.90/98.30 & 99.74/98.93 \\
    \midrule
    \rowcolor{tblblue!25} Average & 97.60/96.70 & 97.50/97.70 & 99.20/98.10 & 98.70/97.80 & 99.05/\underline{98.26} \\
    \midrule
    \rowcolor{tblred!25} Inference time & 159     & 37     & 180   & 28    & \underline{20} \\
    \rowcolor{tblred!25} Parameters  & 97    & 94.7     & 186.55 & 150.64 & \textbf{16.56} \\
    \bottomrule
    \end{tabular}%
  \label{tab:tab1}
\end{table}

\begin{table}
    \caption{Table~\ref{tab:tab1} (continued)}
    \vspace{-\baselineskip}
    \begin{tabular}{lcccccccccccc}
    \toprule
         \; \; Method & SimpleNet~\cite{liu2023simplenet} & DeSTSeg~\cite{zhang2023destseg} & RD++~\cite{tien2023revisiting}  & FOD~\cite{yao2023focus} & Ours \\
    \midrule
    Carpet & \underline{99.70}/98.50 & -/98.30 & \textbf{100}/\underline{99.20} & \textbf{100}/- & 99.48/\textbf{99.91} \\
    Grid  & 99.70/98.80 & -/99.20 & \textbf{100}/\underline{99.30} & \textbf{100}/- & 95.57/95.40 \\
    Leather & \textbf{100}/99.20 & -/\textbf{99.70} & \textbf{100}/99.40 & \textbf{100}/- & \textbf{100}/99.11 \\
    Tile  & 99.80/97.00 & -/99.10 & 99.70/96.60 & \textbf{100}/- & \textbf{100}/\textbf{99.58} \\
    Wood  & \textbf{100}/94.50 & -/\textbf{98.00}  & 99.30/95.80 & 99.1/- & \textbf{100}/\underline{97.99} \\
    Bottle & \textbf{100}/98.00 & -/\textbf{99.40} & \textbf{100}/98.80 & \textbf{100}/- & \textbf{100}/97.77 \\
    Cable & \textbf{99.90}/97.60 & -/97.70 & 99.20/\underline{98.40} & \underline{99.50}/- & 98.98/96.95 \\
    Capsule & 97.70/98.90 & -/\textbf{99.10} & \underline{99.00}/98.80 & \textbf{100.00}/- & 98.69/98.48 \\
    Hazelnut & \textbf{100}/97.90 & -/\textbf{99.80} & \textbf{100}/99.20 &\textbf{100}/- & \textbf{100}/ \underline{99.69} \\
    Metal\_nut & \textbf{100}/98.80 & -/99.00  & \textbf{100}/98.10 & \textbf{100}/- & \textbf{100}/\underline{99.45} \\
    Pill  & \textbf{99.00}/98.60 & -/\textbf{99.10} & 98.40/98.30 & 98.40/- & 98.20/98.61 \\
    Screw & 98.20/99.30 & -/98.80 & \underline{98.90}/\textbf{99.70} & 96.7/- & \textbf{99.47}/97.56 \\
    Toothbrush & 99.70/98.50 & -/99.40 & \textbf{100}/99.10 & 94.4/- & 99.06/\textbf{99.61} \\
    Transistor & \textbf{100}/97.60 & -/92.50 & 98.50/94.30 & \textbf{100.0}/- & 97.58/94.81 \\
    Zipper & \underline{99.90}/98.90 & -/\textbf{99.60} & 98.60/98.80 & 99.70/- & 99.74/98.93 \\
    \midrule
    \rowcolor{tblblue!25} Average & \textbf{99.57}/98.14 & 99.00/98.20 & \underline{99.44}/98.25 & 99.20/\textbf{98.30} & 99.05/\underline{98.26} \\
    \midrule
    \rowcolor{tblred!25} Inference time & 39    & \textbf{18}    & 30    & 103   & \underline{20} \\
    \rowcolor{tblred!25} Parameters  & 52.88 & 35.16 & 154.87 & \underline{28.83}     & \textbf{16.56} \\
    \bottomrule
    \end{tabular}%
    \label{tab:tab15}\\
    \footnotesize{\textbf{Bold} indicates the best and \underline{underline} the second best. All experiments are consistently conducted on an NVIDIA GeForce RTX 3090.}
\end{table}

\subsection{Quantitative Results and Comparison}
\subsubsection{\textbf{Anomaly Detection and Localization on MVTec}}
\label{sec: mvtec}

For the purpose of ensuring fair comparisons with other studies, we use the results reported in the original papers or the results re-evaluated by~\cite{zhang2023destseg,zhang2023prototypical}. In cases where results are not available, we indicate with a hyphen (`-').

Table~\ref{tab:tab1} presents the results of anomaly detection and localization on the MVTec dataset. On average, our method achieves a competitive performance and secures the highest score for 9 out of 15 classes. The specific results in Figure~\ref{fig1} are shown in Table~\ref{tab:tab1}, our method achieves the most efficient method and the second-fastest inference speed, demonstrating the model efficiency and mobile-friendliness of our STLM.

All experiments are consistently conducted on an NVIDIA GeForce RTX 3090. Notably, we attempt to analyze the reason why our STLM has fewer parameters than DeSTSeg but longer inference times is \textbf{the software and hardware optimization works for ResNet-like frameworks are more mature compared to Transformer}~\cite{Li_2023_ICCV}. The optimal inference speed of the image encoder requires further study, which we plan to optimize in future work.

\begin{table}\
\centering
\caption{Anomaly Detection and Localization on MVTec~\cite{bergmann2019mvtec}, VisA~\cite{zou2022spot}, MVTec LOCO~\cite{Loco2022} and DAGM~\cite{zavrtanik2021draem}}
    \begin{tabular}{lcccccccc}
    \toprule
    \multirow{2}{*}{\; \; \; \; \; \, Method} & \multicolumn{4}{c}{VisA}      & \multicolumn{4}{c}{MVTec} \\
    \cmidrule(lr){2-5}\cmidrule(lr){6-9}
    & \multicolumn{1}{c}{I ↑} & \multicolumn{1}{c}{P ↑} & \multicolumn{1}{c}{O ↑} & \multicolumn{1}{c}{A ↑} & \multicolumn{1}{c}{I ↑} & \multicolumn{1}{c}{P ↑} & \multicolumn{1}{c}{O ↑} & \multicolumn{1}{c}{A ↑}\\
    \midrule
    PatchCore~\cite{li2021cutpaste} (186.6M) &95.10 & \textbf{98.80} & 91.20 & 40.10 & 99.20 & 98.10 & 93.40 & 56.10 \\
    RD4AD~\cite{roth2022towards} (150.6M) &96.00 & 90.10 & 70.90 & 27.70 & 98.70 & 97.80 & 93.93 & 58.00\\
    RD++~\cite{tien2023revisiting} (154.9M) & 95.90 & 98.70 & 93.40 & 40.80 & 99.44 & 98.25 & 94.99 & 60.80 \\
    DeSTSeg~\cite{zhang2023destseg} (35.2M) & 90.02 & 40.76 & 99.00 & 98.20 & \textbf{95.11} & 75.80 & 89.72 & 57.85 \\
    SimpleNet~\cite{liu2023simplenet} (52.9M) & 88.70 & 36.30 & \textbf{99.57} & 98.14 & 90.00 & 54.80 & 82.65 & 46.40 \\
    FastFlow~\cite{yu2021fastflow} (92M) & 59.80 & 15.60 & 90.50 & 95.50 & 85.60 & 39.80 & 77.38 & 29.87 \\
    DRAEM~\cite{zavrtanik2021draem} (97M)  & 73.70 & 30.50 & 97.60 & 96.70 & 92.10 & 68.40 & 74.40 & 43.17 \\
    \rowcolor[gray]{.85}  Ours (\textbf{16.6M}) & \textbf{93.83} & \textbf{47.63} & 99.05 & \textbf{98.26} & 94.92 & \textbf{76.32} &  \textbf{93.18} & \textbf{90.51}  \\
    \bottomrule
    \end{tabular}%
    \label{tab:tab3}
\end{table}

\begin{table}
\centering
\caption{Table~\ref{tab:tab3} (continued)}
    \begin{tabular}{lcccccccccc}
    \toprule
    \multirow{2}{*}{\; Method} & \multicolumn{2}{c}{MVTec LOCO} & \multicolumn{4}{c}{DAGM}     & \multicolumn{4}{c}{Average} \\
   \cmidrule(lr){2-3}\cmidrule(lr){4-7}\cmidrule(lr){8-11}
     &\multicolumn{1}{c}{I ↑} & \multicolumn{1}{c}{O ↑ }& \multicolumn{1}{c}{I ↑} & \multicolumn{1}{c}{P ↑} & \multicolumn{1}{c}{O ↑} & \multicolumn{1}{c}{A ↑} & \multicolumn{1}{c}{I ↑} & \multicolumn{1}{c}{P ↑} & \multicolumn{1}{l}{O ↑} & \multicolumn{1}{l}{A ↑} \\
    \midrule
     PatchCore & 84.80 & 64.30& 93.60 & 96.70 & 89.30 & 51.70 & 93.18 & 97.87 & 84.55 & 49.30 \\
     RD4AD & 88.00 & 84.80 & 95.80 & \textbf{97.50} & 93.00 & 53.40 & 94.63 & 95.13 & 85.66 & 46.37 \\
     RD++ & 92.45 & 55.30 & \textbf{98.50} & 97.40 & \textbf{93.80} & 64.30 & 95.41 & \textbf{98.12} & 92.45 & 55.30 \\
    DeSTSeg &91.95 & 97.73  & 97.44 & 93.55 & 87.88 & 56.99  & 94.71 & 96.49 & 89.72 & 57.85 \\
    SimpleNet &\textbf{96.80} & 97.80 & 95.30 & 97.10 & 91.30 & 48.10 & 93.84 & 97.68 & 82.65 & 46.40 \\
    FastFlow &82.20 & 88.20 & 87.40 & 91.10 & 79.90 & 34.20 & 85.75 & 91.60 & 77.38 & 29.87 \\
    DRAEM &88.70 & 94.40 & 90.80 & 86.80 & 71.00 & 30.60 & 87.88 & 92.63 & 74.40 & 43.17 \\
    \rowcolor[gray]{.85}  Ours&96.73 & 98.36 & 98.30 & 96.33 & 91.14 & \textbf{64.91} & \textbf{96.81} & 97.65 & \textbf{92.60} & \textbf{62.95} \\
    \bottomrule
    \end{tabular}%
    \label{tab:tab16}\\
    \footnotesize{``I'', ``P'', ``O'' and ``A'' respectively refer to the five metrics of image-level AUROC, pixel-level AUROC, PRO and AP. The best results are highlighted in \textbf{bold}.}
\end{table}

\begin{table}
      \caption{Additional experiments on model parameters and inference flexibility.}
    \begin{tabular}{lcccc}
    \toprule
    \; \; \; Method & \makecell{Inference time\\ (ms)} & \makecell{Parameter\\($\times 10^6$)} & \makecell{ GFLOPS\\($\times 10^9$)} & \makecell{ GPU Memory\\(MiB)} \\
    \midrule
    DRAEM~\cite{zavrtanik2021draem} & 159   & 97.4  & 696   & - \\
    FastFlow~\cite{yu2021fastflow} & 43  & 92.18 & 108   & 6348 \\
    PatchCore~\cite{roth2022towards} & 180 & 186.55 & 66+kNN & 10010+kNN \\
    RD4AD~\cite{deng2022anomaly} & 28  & 150.64 & \textbf{39} & 3219 \\
    SimpleNet~\cite{liu2023simplenet} & 39  & 52.88 & 57    & 7983 \\
    DeSTSeg~\cite{zhang2023destseg} & \textbf{18} & 35.16 & 40    & 2527 \\
    RD++~\cite{tien2023revisiting}  & 30  & 166.10 & 57    & 4761 \\
    \rowcolor[gray]{.85}STLM (Ours)  & 20  & \textbf{16.56} & 55    & \textbf{2005} \\
    \bottomrule
    \end{tabular}%
  \label{tab:tab12}
\end{table}

\begin{figure}%
	\centering
\includegraphics[width=1\columnwidth]{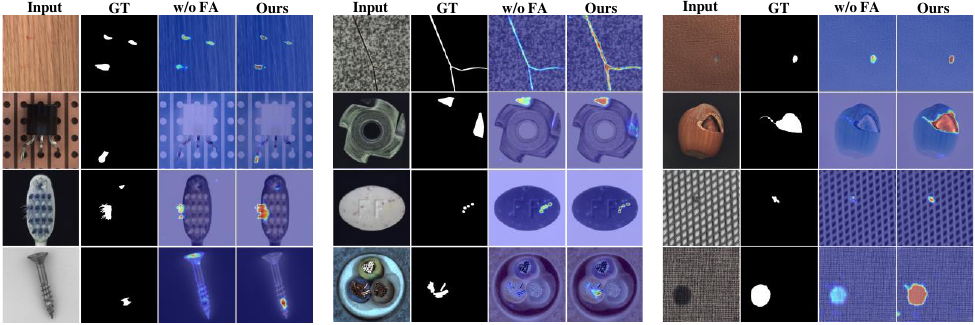}
\caption{Visualization examples of our method on MVTec. The Feature Aggregation (FA) module is always effective.}
	\label{fig6}
\end{figure}

\begin{figure}%
	\centering
\includegraphics[width=1\columnwidth]{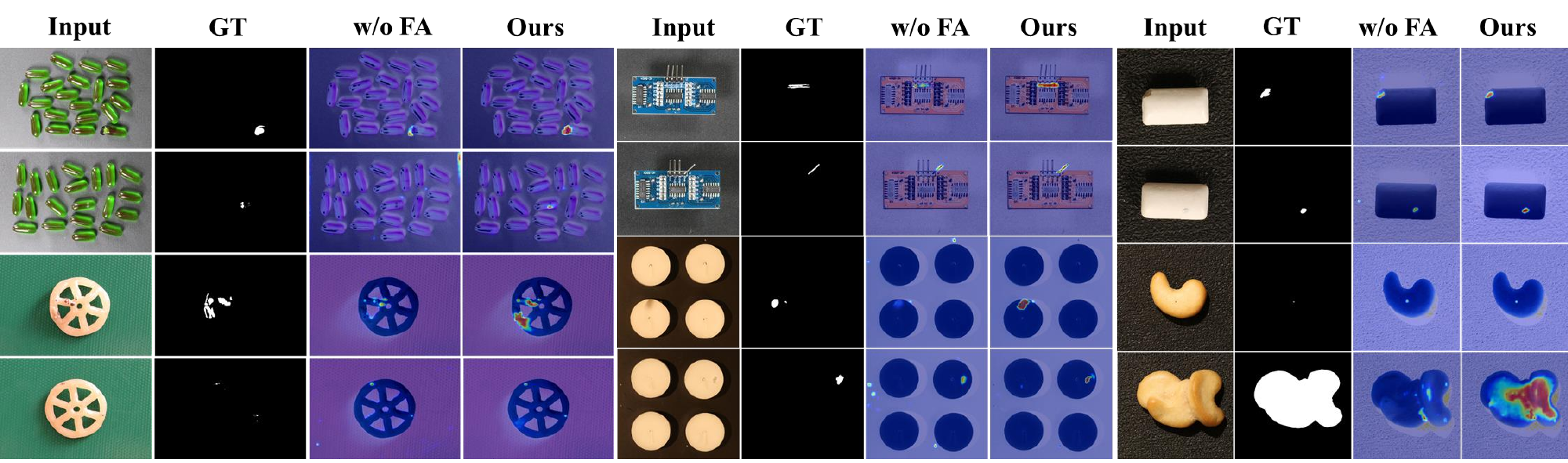}
\caption{Visualization examples of our method on VisA.}
	\label{fig7}
\end{figure}

\begin{wrapfigure}{R}{0.6\textwidth} 
\includegraphics[width=0.6\columnwidth]{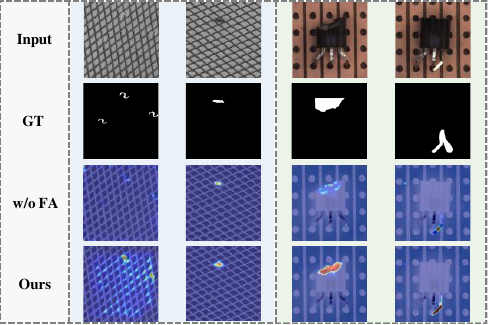}
\caption{Failure cases of our method. The examples are selected from grid and transistor (from left to right).}
	\label{fig8}
\end{wrapfigure}

\subsubsection{\textbf{Evaluations on other more difficult benchmarks}}
\label{sec: DAGM}
To conduct more comprehensive evaluations of the anomaly detection capabilities, we further subject our networks to benchmarking using three additional widely recognized datasets VisA~\cite{zou2022spot}, MVTec LOCO~\cite{Loco2022} and DAGM~\cite{zavrtanik2021draem}. VisA features different objects in more types, incorporating three diverse types: complex structures, multiple instances, and single instances. MVTec LOCO contains images from five different categories inspired by real-world industrial inspection scenarios. Structural anomalies appear as scratches, dents, or contaminations in the manufactured products. DAGM contains 10 textured objects with smaller abnormal regions that bear a strong visual resemblance to the background. To demonstrate the intricacy of these three datasets, we present a comparison with the MVTec dataset in Figure~\ref{fig13}. Furthermore, the results across all methods reveal a consistent decline for these three datasets.
 
To provide a comprehensive overview of anomaly localization capabilities, we include 4 metrics: image-level AUROC, pixel-level AUROC, PRO and AP.
The results, as presented in Table~\ref{tab:tab3}, indicate that our method achieves SOTA pixel-level PRO and AP metrics of 93.83/47.63 on the VisA dataset and SOTA image-level AUROC and pixel-level PRO metrics of 93.18/90.51 on the VisA dataset. Our method has also demonstrated its superiority on the DAGM dataset. On average, STLM performs exceptionally well, affirming its effectiveness and scalability when faced with increasingly complex datasets.

\subsubsection{\textbf{Evaluations on model parameters and inference flexibility}}
\label{sec: MPIF}
We have conducted additional experiments regarding model parameters and inference flexibility, comparing our method with those mentioned in our paper. The specific results are presented in ~\ref{tab:tab12}. Overall, our method achieves superior results compared to other methods on average across four datasets. Additionally, our method demonstrates significant advantages in terms of the number of parameters and GPU memory usage, which is highly valuable for practical industrial applications. Moreover, our method achieves the second-fastest speed and ranks third in GFLOPS, making it the best overall in terms of both performance and efficiency.

\subsection{Qualitative Results and Comparison}
\subsubsection{\textbf{Visualization on MVTec}}
\label{sec: vis}
To further analyze the model performance, we visualize a qualitative evaluation of anomaly localization performance presented in Figure~\ref{fig6}. The visual results demonstrate our model's precise localization of anomalies. However, no one is perfect. In Figure~\ref{fig8}, we analyze instances where our method does not perform as expected. As evidenced in the first and second columns, the susceptibility of our FA module holds the primary responsibility. But as mentioned in~\cite{zhang2023destseg}, we must acknowledge that several uncertain ground truths are accountable for the observed failures. Considering the results in the third and fourth columns where the ground truth highlights the designated location, our masks only cover the anomalous region on the same images.

\subsubsection{\textbf{Visualization on other benchmarks}}
\label{sec: visOther}
We conducted substantial qualitative experiments on VisA datasets to visually demonstrate the superiority of our method in the accuracy of anomaly localization on other more difficult benchmarks. Also, the Feature Aggregation (FA) module is always effective. As shown in Figure~\ref{fig7}, the visual results demonstrate our model's robust generalization. 

\subsection{Ablation Studies}
\label{sec: ablation}

\subsubsection{\textbf{The importance of Large Teacher, Plain Student Stream, Mask Decoder, and Feature Aggregation}}
In Table~\ref{tab:tab4}, we assess the effectiveness of our four design components by conducting experiments where we eliminate the fixed teacher network and use a pre-trained mobileSAM as the teacher network to examine the contribution of SAM-knowledge, remove the plain student stream and use the fixed SAM as the plain stream in the TLM like early KD frameworks (but still use both normal and pseudo anomaly images to train denoising student stream), employee the image embedding exacted from the encoders as the FA module input and train with the cosine loss between paired outputs of two encoders, and substitute the FA module with an empirical feature fusion strategy~\cite{wang2021student}. The best results are achieved when all key design components are combined. 

Notably, our work outperforms the ``w/o-PlainS'' manner, explaining that the plain student stream can learn generalized knowledge related to anomaly detection tasks. Based on the results from Section~\ref{sec: oneStage}, where one-stage training outperforms two-stage training, we believe this is due to the FA and TLM modules being trained simultaneously. The segmentation knowledge related to anomaly detection images is transmitted forward by the loss of the FA module, helping both students better recognize normal regions. Without distillation training for SAM, only the denoising student is improved, resulting in differences in normal regions and affecting the experimental results.This enables our two student streams to effectively represent previously unseen patterns, ensuring differences between features of two streams are well-captured.

\begin{table}
      \caption{Ablation studies on our main designs on the MVTec dataset~\cite{bergmann2019mvtec}.}
  \begin{tabular}{ccccccc}
    \toprule
    Large Teacher & Plain Student & Mask Decoder & FA & I-AUROC  & P-AUROC  & PRO  \\
    \midrule
    Mobile Teacher & \checkmark  & \checkmark & \checkmark & 96.36 & 97.57 & 93.07 \\
    \checkmark & - & \checkmark & \checkmark & 98.03 & 95.77 & 89.96 \\
    \checkmark & \checkmark & - & \checkmark & 97.74 & 95.65 & 88.05 \\
    \checkmark & \checkmark & \checkmark &  \cite{wang2021student} & 93.68 & 96.81 & 91.58 \\
    \rowcolor[gray]{.85}  \checkmark & \checkmark & \checkmark & \checkmark & \textbf{99.05} & \textbf{98.26} & \textbf{94.92} \\
    \bottomrule
  \end{tabular}
  \label{tab:tab4}
   \footnotesize{I-AUROC, P-AUROC, and PRO ($\%$) are used to evaluate image-level and pixel-level detection. The best results are highlighted in \textbf{bold}.}
\end{table}

\begin{table}
\parbox{.47\linewidth}{
   \caption{Ablation studies on training strategy between one-stage and two-stage training.}
    \begin{tabular}{lccc}
    \toprule
    \;\;Method & I-AUROC  & P-AUROC  & PRO  \\
    \midrule
    \rowcolor[gray]{.85}  One-stage & \textbf{99.05}  & \textbf{98.26}  & \textbf{94.92}  \\
    Two-stage & 98.98  & 98.06  & 90.12  \\
    \bottomrule
     \label{tab:tab5}  
  \end{tabular}}
  \hfill
\parbox{.47\linewidth}{
        \caption{Ablation studies on the input of feature aggregation module on the MVTec dataset~\cite{bergmann2019mvtec}.}
  \begin{tabular}{lccc}
    \toprule
    \;Method & I-AUROC  & P-AUROC  & PRO  \\
    \midrule
    Cosine & 98.95  & 94.61  & 92.63  \\
    Sub & 99.00  & 94.30  & 93.96  \\
    Concat & 98.89  & 97.98  & 94.51  \\
    \rowcolor[gray]{.85} Ours  & \textbf{99.05}  & \textbf{98.26}  & \textbf{94.92}  \\
    \bottomrule
      \label{tab:tab6}
  \end{tabular}}
\end{table}

\subsubsection{\textbf{Effect for One-stage training strategy}}
\label{sec: oneStage}
We conduct ablation studies to investigate the impact of two distinct training strategies mentioned in Section~\ref{sec: FA}, as detailed in Table~\ref{tab:tab5}. The results reveal that one-stage training delivered superior performance compared to separate training of the TLM and FA module proposed in~\cite{zhang2023destseg}, particularly in terms of PRO scores.  This suggests that training the TLM with additional supervised signals of the pseudo anomaly mask $M$ enhances the network's ability to locate tiny defects or accurately delineate defect boundaries.

\subsubsection{\textbf{Discrepancy among different manners for the input of the Feature Aggregation module}}
As mentioned in Section~\ref{sec: FA}, the input of our feature aggregation module is two element-wise products of the feature maps from TLM, as defined by Equation~\eqref{eqn-2}. To validate the rationality of this setting, we evaluate three alternative feature combinations as input.

The first way computes the cosine distance between the feature maps of our TLM, making use of more prior information from STLM, which is trained through optimization of the cosine loss function. The second way involves concatenating features of the normal image and the anomalous residual representation defined by~\cite{zhang2023prototypical}. The anomalous feature residuals denote the element-wise Euclidean distance between tensors from two streams. The third way directly concatenates the feature maps $S_{P}^{k}$ and $S_{D}^{k}$ as the input of our FA module. The second and the third ways preserve the information from TLM more effectively. The results are presented in Table~\ref{tab:tab6}, indicating our method maintains a balance between prior knowledge and representation of information.


\begin{figure}
    \centering
    \subfigure[Comparison between decoders]{\includegraphics[width=0.45\columnwidth]{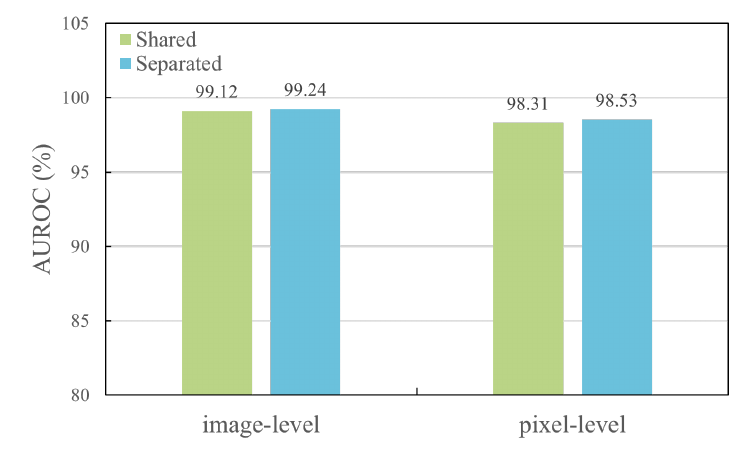}\label{fig4_a}}
    \subfigure[Comparison between FA module]{\includegraphics[width=0.45\columnwidth]{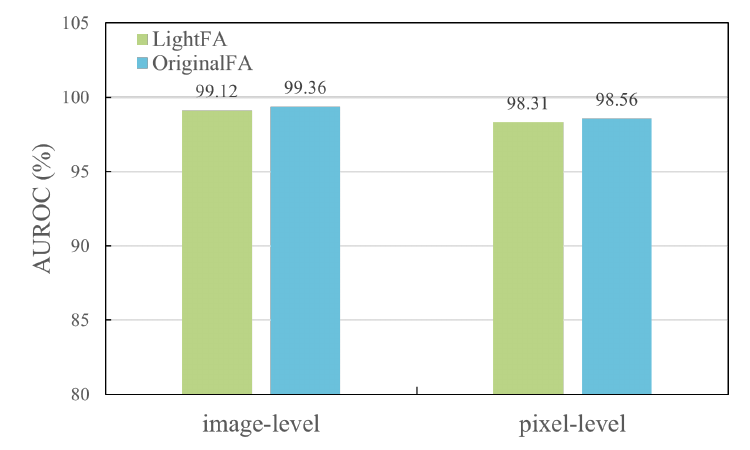}\label{fig4_b}}
    \subfigure[the probability of being activated]{\includegraphics[width=0.45\columnwidth]{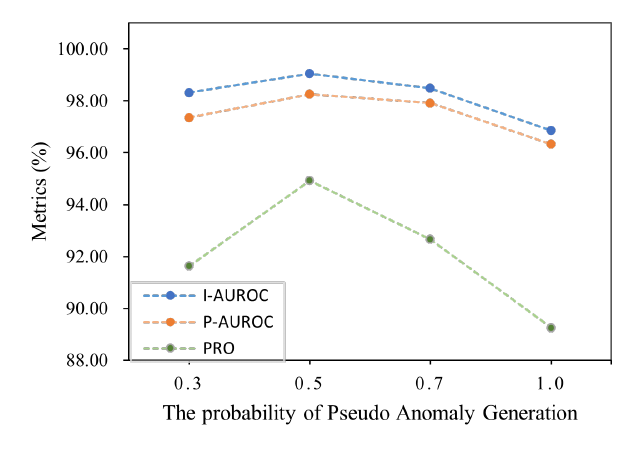}\label{fig4_c}}
    \subfigure[the $k$-th feature from mask decoder]{\includegraphics[width=0.45\columnwidth]{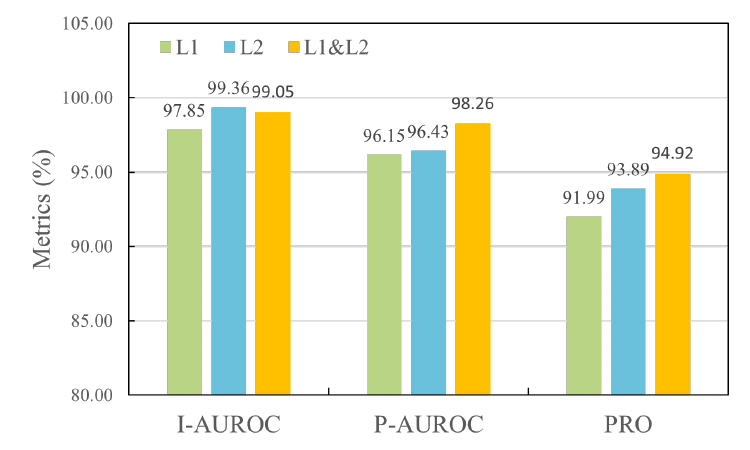}\label{fig4_d}}
    \caption{(a) Comparison between one shared decoder and two separated decoders. (b) Comparison between lightweight and original feature aggregation module. (c) Experience on the probability of Pseudo Anomaly Generation being activated. (d) Ablation studies on the feature extracted from the $k$-th layer of mask decoder.} 
    \label{fig4}
\end{figure}

\subsubsection{\textbf{Ablation studies on the decoder and feature aggregation module}} The image embeddings generated by the student encoder can closely approximate that of the original teacher encoder, leading to the conclusion that the utilization of distinct decoders for the TLM may not be imperative. as shown in Figure~\ref{fig4_a}, We observe that the shared decoder, while economical in terms of parameters and computation time, achieves a satisfactory result with a slight decrease compared to employing two separated decoders. Furthermore, as indicated in Figure~\ref{fig4_b}, it has been observed that reducing the depth of the FA module does not significantly impact the performance. As a result, an effort has been made to make this module more lightweight by reducing the channels from 256 to 128 and adjusting the dilation rate to [1, 1, 3], following a similar method in~\cite{zhang2023cross}.  These two lightweight steps mitigate memory costs associated with both training and inference, which is a factor of paramount importance in practical implementations. 

\subsubsection{\textbf{Analysis of the probability of Pseudo Anomaly Generation}} The probability of the Pseudo Anomaly Generation process being activated governs the extent to which the training datasets deviate from the original normal datasets. To be specific, a higher probability results in a larger proportion of abnormal images and a smaller proportion of normal images. The probability of 0.5 denotes an optimal balance between pseudo-anomalous images and normal images, as shown in Figure~\ref{fig4_c}, achieving the best performance. Since the ``good'' samples in the test dataset, the probability of 1.0 lets normal images be ``unseen'' images during training, leading to a high false negative. We design experiments to verify this explanation in Section~\ref{sec: FNR}.


\subsubsection{\textbf{Ablation studies on the feature extracted from the $k$-th layer of mask decoder}}   In Equation~\eqref{eqn-3} and Equation~\eqref{eqn-4}, we define the layers of the feature extracted from the Mask Decoder as the $k$-th layer, which is also presented in Figure~\ref{fig2} (d): Element-wise Production. We conducted a series of experiments to investigate the impact of $k$, and the corresponding results are presented in Figure~\ref{fig4_d}. We denoted the two layers as L1 and L2. We include image-level AUROC, pixel-level AUROC and PRO metrics to provide a comprehensive cooperation of the anomaly detection and localization capabilities of each layer. 

It is evident that features extracted from L1 alone already yield excellent image-level AUROC performance, while pixel-level AUROC and PRO metrics benefit from incorporating information from both L1 and L2. It indicates that it is necessary to exchange information across different layers. Therefore, we selected the combination of L1 + L2 as our default setting.


\begin{figure}%
	\centering
\includegraphics[width=0.9\columnwidth]{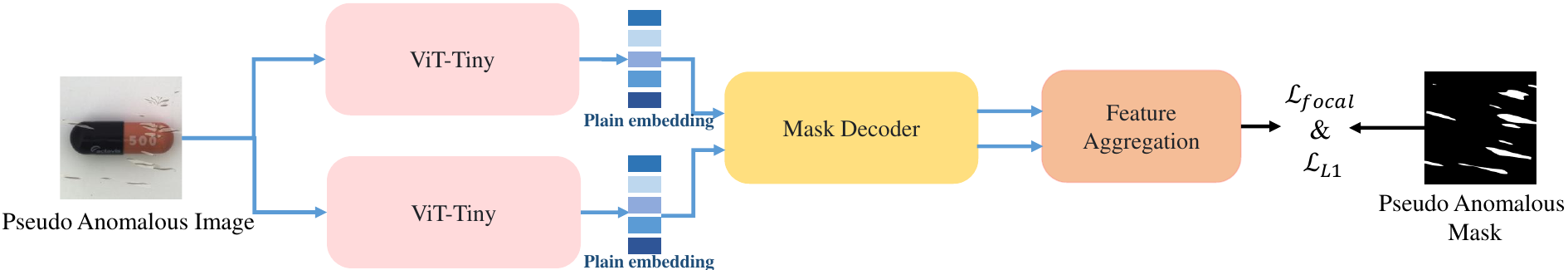}
\caption{Overview of the ``w/o fixed SAM'' method. TLM is used during training and inference.}
	\label{fig11}
\end{figure}
\begin{figure}%
	\centering
\includegraphics[width=0.9\columnwidth]{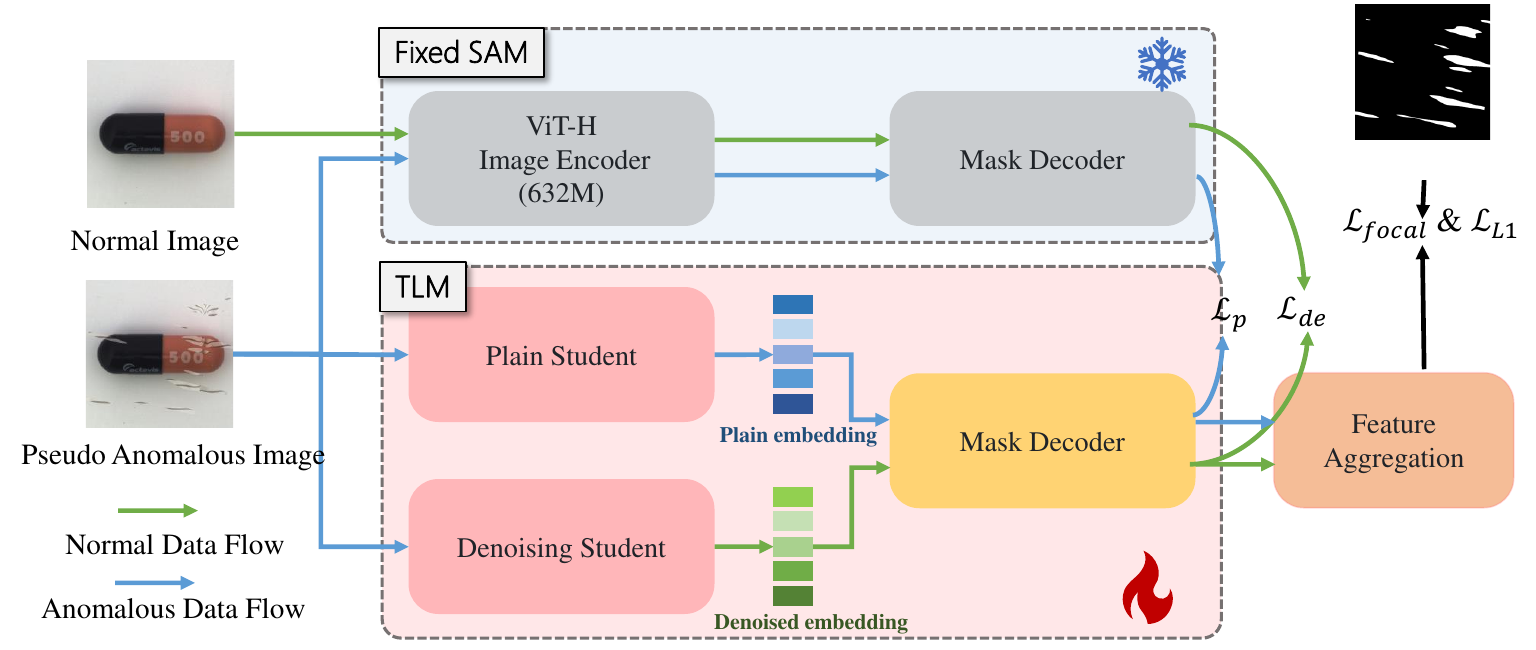}
\caption{Overview of the logit-based method. One stream is trained to segment both normal and anomalous regions as the SAM does, while the other generates the mask without anomalies. For \textbf{inference}, the architecture is the same as STLM.}
	\label{fig12}
\end{figure}

\subsubsection{\textbf{Effect of the knowledge distillation framework}} We investigate the importance of the fixed SAM teacher, \ie, only utilizing TLM during training and inference phases, as illustrated in Figure~\ref{fig11}.In this scenario, we no longer use the knowledge distillation structure. Instead, we directly train TLM like a regular segmentation network, using the mask as a supervision signal to segment anomalous regions. The results, shown in Table~\ref{tab:tab7}, indicate that directly using a segmentation network to identify anomalous regions does not yield satisfactory results. In contrast, using a distillation model for differential learning tasks in TLM's two students produces better outcomes. 

\subsubsection{\textbf{Ablation studies on the distillation methods}} SAM generates high-quality segmentation results for previously unseen images, making it suitable for logit distillation.
As \cite{zhao2022decoupled} said, distillation methods often focus on distilling deep features from intermediate layers, while the significance of logit distillation is greatly overlooked. We compare the feature-based method with the logit-based method, which minimizes the focal loss and L1 loss between segmentation masks of teacher SAM and the student streams, as illustrated in Figure ~\ref{fig12}. We no longer use feature distillation; instead, we attempt to leverage SAM's logit distillation capabilities. In this experiment, we directly use the segmentation results from SAM for distillation, aligning TLM's segmentation results with the fixed SAM teacher. The experimental results, shown in Table~\ref{tab:tab8}, indicate that the performance of feature distillation is superior on average for our model, compared with directly utilizing segmentation results from a SAM-style architecture.

\begin{table}
\parbox{.47\linewidth}{
  \centering
        \caption{Ablation studies on the knowledge distillation framework: I-AUROC, P-AUROC, and PRO ($\%$) are used to evaluate image-level and pixel-level detection on the MVTec dataset \cite{bergmann2019mvtec}.}
  \begin{tabular}{lccc}
    \toprule
    \; Method & I-AUROC  & P-AUROC  & PRO  \\
    \midrule
    w/o SAM & 89.40   &  90.02 &  79.19  \\
    \rowcolor[gray]{.85}  Ours  & \textbf{99.05}  & \textbf{98.26}  & \textbf{94.92}  \\
    \bottomrule
  \end{tabular}
  \label{tab:tab7}}
  \hfill
\parbox{.47\linewidth}{
  \centering
        \caption{Ablation studies on the distillation methods: I-AUROC, P-AUROC, and PRO ($\%$) are used to evaluate image-level and pixel-level detection on the MVTec dataset \cite{bergmann2019mvtec}.}
  \begin{tabular}{lccc}
    \toprule
    \;\; Method & I-AUROC  & P-AUROC  & PRO  \\
    \midrule
    Logit & 93.45   &  91.65  &  80.38  \\
    \rowcolor[gray]{.85} Ours  & \textbf{99.05}  & \textbf{98.26}  & \textbf{94.92}  \\
    \bottomrule
  \end{tabular}
  \label{tab:tab8}}
\end{table}

\subsubsection{\textbf{Additional experience of the probability of Pseudo Anomaly Generation.}}
\label{sec: FNR}
In Table~\ref{tab:tab10}, we provide further quantitative results of the false negative rate (\textit{FNR}) \cite{altman1994statistics} to verify the analysis in the ablation study that the probability of 1.0 lets normal images be ``unseen'' images during training, leading to a high false negative. Furthermore, the TLM and FA are trained under the assumption that all pictures are abnormal, causing the model to "think" all pictures are anomalies and incorrectly classify normal images as abnormal, thereby reducing performance. The false negative rate, \ie, miss rate, is defined to measure the proportion of instances that are incorrectly identified as negative when it is actually positive. The \textit{FNR} is calculated as:
\begin{equation}\label{eqn-9} 
  FNR = \frac{FN}{TP + FN},
\end{equation}
where false negative (\textit{FN}) refers to the number that is actually positive but incorrectly predicted as negative and true positive (\textit{TP}) represents the total number that is actually positive.

\begin{table}
\parbox{.47\linewidth}{
\caption{Additional experience of the probability of Pseudo Anomaly Generation: FNR ($\%$) on the MVTec dataset \cite{bergmann2019mvtec}.}
    \begin{tabular}{lcc|c}
    \toprule
    \;\;\,Method & Avg. Text.↓ & Avg. Obj.↓  & Avg.↓ \\
    \midrule
    Prob = 1.0  &\textbf{ 35.77}   &  38.59 & 37.71  \\
    \rowcolor[gray]{.85}  Prob = 0.5  & 35.87   &  \textbf{30.64}   &   \textbf{32.39}  \\
    \bottomrule
    \end{tabular}%
  \label{tab:tab10}}
  \hfill
\parbox{.47\linewidth}{
  \caption{Ablation studies on different anomaly generation strategies on the MVTec dataset~\cite{bergmann2019mvtec}.}
  \begin{tabular}{lccc}
      \toprule
    \;Method & I-AUROC  & P-AUROC  & PRO  \\
        \midrule
    \rowcolor[gray]{.85}  NSA   & \textbf{99.05} & \textbf{98.26} & \textbf{94.92} \\
    Parch & 84.74 & 89.85 & 82.03 \\
    Scar & 96.63 & 96.30 & 90.52 \\
    Union & 97.49 & 97.16 & 91.06 \\
    \bottomrule
  \end{tabular}
  \label{tab:tab11}}
\end{table}

\subsubsection{\textbf{Evaluation on different anomaly generation strategies.}}
To evaluate the pseudo anomaly generation strategies, we present experimental results with the renowned strategies NSA~\cite{NSA2022} and Cutpaste~\cite{li2021cutpaste} in Table~\ref{tab:tab11}. In most SOTA methods, \ie, DRAEM~\cite{zavrtanik2021draem}, Prototypical~\cite{zhang2023prototypical} and DeSTSeg~\cite{zhang2023destseg}, the authors attempt to leverage synthetic anomalies generated by NSA, which are also utilized in our paper. Furthermore, based on NSA and Cutpaste, ~\cite{kim2024sanflow,yao2023explicit,zhang2023prototypical,tien2023revisiting} employ a variety of complex or artificial operations that more closely resemble the real anomalies. Notably, SimpleNet~\cite{liu2023simplenet} simulates anomalies by introducing Gaussian noise to normal features, a process that requires careful tuning of appropriate hyperparameters.

\subsection{Discussion and Future Work}
\label{sec: discussion}
Recent studies in anomaly detection have increasingly focused on vision foundation models, which are more effective than CNNs due to their ability to handle long-range dependencies. Our STLM also leverages vision foundation models but avoids the quadratic computational complexity of transformers. As shown in Table~\ref{tab:tab3}, STLM achieves performance that is highly competitive with state-of-the-art methods, without the long inference time as FOD~\cite{yao2023focus} and the large model parameters as RD++~\cite{tien2023revisiting}. Specifically, our method outperforms other methods on average across four datasets by a notable margin of 5.1$\%$ on AP, 1.4$\%$ on image-level AUROC, and 0.15$\%$ on pixel-level PRO. On pixel-level AUROC, our STLM is only 0.47$\%$ behind the best-performing method RD++. However, as noted by ~\cite{tao2022deep}, Pixel-AUROC may not accurately reflect localization accuracy, since the large number of anomalous-free pixels heavily influences the false positive rate. RD++ employs an additional self-supervised optimal transport loss to suppress anomalous signals before forwarding to the student. It improves the localization accuracy of anomalous-free pixels but not anomalous ones, evidenced by RD++'s extremely high Pixel-AUROC but relatively average PRO.

Additionally, Table~\ref{tab:tab12} shows that our model excels in model parameters and inference flexibility. Large models with high computational demands are unsuitable for large-scale deployment, especially on robots or mobile devices, where slow inference speed also impacts production progress. Our method delivers superior results on average across four datasets while offering significant advantages in parameter efficiency and GPU memory usage, making it highly suitable for robotic and mobile device applications. Moreover, our STLM achieves the second-fastest inference speed and ranks third in GFLOPS. We attribute the discrepancy of STLM between fewer model parameters and longer inference time compared to DeSTSeg to the more mature software and hardware optimization for ResNet-like frameworks over Transformer-based architectures. It is a challenge that requires further study and could lead to performance improvements. In the future, we aim to optimize the inference speed of the image encoder and develop a model for multi-class anomaly detection using a simpler structure.

\section{Conclusion}
\label{sec:conclu}

In industrial anomaly detection, the primary concerns in real-world applications include two aspects: model efficiency and mobile-friendliness. In this paper, we propose a novel framework called SAM-guided Two-stream Lightweight Model for unsupervised anomaly detection tasks, tailored to meet the demands of real-world industrial applications while capitalizing on the strong generalization ability of SAM. Our STLM effectively distills distinct knowledge from SAM into the Two-stream Lightweight Model (TLM), assigning different tasks to each stream. To be specific, one stream focuses on generating discriminative and generalized feature representations in both normal and anomalous locations, while the other stream reconstructs features without anomalies. Last, we employ a shared mask decoder and a feature aggregation module to generate anomaly maps. By the design of our Two-stream Lightweight Model, shared mask decoder and lightweight feature aggregation module, the calculation and the number of parameters of the networks can be significantly reduced and the inference speed can naturally be improved without noticeable accuracy loss. Experiments on the renowned four datasets demonstrate that our method achieves competitive results on anomaly detection tasks. Furthermore, it exhibits model efficiency and mobile-friendliness compared to state-of-the-art methods.


\bibliographystyle{ACM-Reference-Format}
\bibliography{bibliography}

\end{document}